\documentclass[conference]{IEEEtran}
\IEEEoverridecommandlockouts
\usepackage{cite}
\usepackage{amsmath,amssymb,amsfonts}
\usepackage{algorithmic}
\usepackage{graphicx}
\usepackage{textcomp}
\usepackage{xcolor}
\usepackage{xspace}
\usepackage{comment}
\usepackage{multicol, multirow}
\usepackage{caption}
\usepackage{subcaption}
\usepackage{booktabs}       
\def\BibTeX{{\rm B\kern-.05em{\sc i\kern-.025em b}\kern-.08em
    T\kern-.1667em\lower.7ex\hbox{E}\kern-.125emX}}
\begin{document}

\newcommand{\zd}[1]{{\color{blue}{\it DZ: #1}}}
\newcommand{\sys}{DistDGL\xspace}

\title{\sys: Distributed Graph Neural Network Training for Billion-Scale Graphs
}

\author{\IEEEauthorblockN{Da Zheng}
\IEEEauthorblockA{\textit{AWS AI}\\
dzzhen@amazon.com}
\and
\IEEEauthorblockN{Chao Ma}
\IEEEauthorblockA{\textit{AWS Shanghai AI Lab} \\
manchao@amazon.com}
\and
\IEEEauthorblockN{Minjie Wang}
\IEEEauthorblockA{\textit{AWS Shanghai AI Lab} \\
minjiw@amazon.com}
\and
\IEEEauthorblockN{Jinjing Zhou}
\IEEEauthorblockA{\textit{AWS Shanghai AI Lab} \\
zhoujinj@amazon.com}
\and
\IEEEauthorblockN{Qidong Su}
\IEEEauthorblockA{\textit{AWS Shanghai AI Lab} \\
qidos@amazon.com}
\and
\IEEEauthorblockN{Xiang Song}
\IEEEauthorblockA{\textit{AWS Shanghai AI Lab} \\
xiangsx@amazon.com}
\and
\IEEEauthorblockN{Quan Gan}
\IEEEauthorblockA{\textit{AWS Shanghai AI Lab} \\
quagan@amazon.com}
\and
\IEEEauthorblockN{Zheng Zhang}
\IEEEauthorblockA{\textit{AWS Shanghai AI Lab} \\
zhaz@amazon.com}
\and
\IEEEauthorblockN{George Karypis}
\IEEEauthorblockA{\textit{AWS AI} \\
gkarypis@amazon.com}
}

\maketitle

\begin{abstract}
Graph neural networks (GNN) have shown great success in learning from graph-structured data.
They are widely used in various applications, such as recommendation, fraud detection,
and search. In these domains, the graphs are typically large, containing hundreds of millions
of nodes and several billions of edges. To tackle this challenge, we develop \sys, a system
for training GNNs in a mini-batch fashion on a cluster of machines. \sys is based on the
Deep Graph Library (DGL), a popular GNN development framework.
\sys distributes the graph and its associated data (initial features and embeddings) across
the machines and uses this distribution to derive a computational decomposition by following
an owner-compute rule. \sys follows a synchronous training approach and allows ego-networks
forming the mini-batches to include non-local nodes. To minimize the overheads associated
with distributed computations, \sys uses a high-quality and light-weight min-cut graph
partitioning algorithm along with multiple balancing constraints. This allows it to reduce
communication overheads and statically balance the computations. It further reduces the
communication by replicating halo nodes and by using sparse embedding updates.
The combination of these design choices allows \sys to train high-quality models while
achieving high parallel efficiency and memory scalability.
%
%
We demonstrate our optimizations on both inductive and transductive GNN models.
Our results show that \sys achieves linear speedup without compromising model
accuracy and requires only 13 seconds to
complete a training epoch for a graph with 100 million nodes and 3 billion edges on
a cluster with 16 machines. DistDGL is now publicly available as part of DGL: https://github.com/dmlc/dgl/tree/master/python/dgl/distributed.

\end{abstract}

\begin{IEEEkeywords}

\end{IEEEkeywords}

\section{Introduction}

Graph Neural Networks (GNNs) have shown success in learning from graph-structured data and have
been applied to many graph applications in social networks, recommendation,
knowledge graphs, etc. In these applications, graphs are usually huge, in the order of
many millions of nodes or even billions of nodes. For instance, Facebook social network
graph contains billions of nodes. Amazon is selling billions of items and has billions of
users, which forms a giant bipartite graph for its recommendation task.
Natural language processing tasks take advantage of knowledge graphs,
such as Freebase \cite{freebase} with 1.9 billion triples.

It is challenging to train a GNN model on a large graph.
Unlike domains such as computer vision and natural language processing, where training
samples are mutually independent, graph inherently represents the dependencies among training
samples (i.e., vertices). Hence, mini-batch training on GNNs is different
from the traditional deep neural networks; each mini-batch must incorporate
those depending samples. The number of depending samples usually grows exponentially
when exploring more hops of neighbors. 
This leads to many efforts in designing various sampling algorithms to scale GNNs to
large graphs \cite{graphsage, chen2017stochastic, fastgcn, huang2018, pinsage}.
The goal of these methods is to prune the vertex dependency to reduce
the computation while still estimating the vertex representation computed
by GNN models accurately.

It gets even more challenging to train GNNs on giant graphs when scaling beyond
a single machine. For instance, a graph with billions of nodes requires memory
in the order of terabytes attributing to large vertex features and edge features.
Due to the vertex dependency, distributed GNN training requires to
read hundreds of neighbor vertex data to compute a single vertex representation,
which accounts for majority of network traffic in distributed GNN training.
This is different from traditional distributed neural network training,
in which majority of network traffic comes from exchanging the gradients of
model parameters. In addition, neural network models are typically trained
with synchronized stochastic gradient
descent (SGD) to achieve good model accuracy. This requires the distributed GNN
framework to generate balanced mini-batches that contain roughly the same number of
nodes and edges as well as reading the same account of data from the network.
Due to the complex subgraph structures in natural graphs, it is difficult to
generate such balanced mini-batches.

Unfortunately, current systems cannot effectively address the challenges of
distributed GNN training.
Previous distributed graph analytical systems~\cite{pregel,powergraph,ligra}
are designed for full graph computation expressed in the vertex-centric program paradigm,
which is not suitable for GNN mini-batch training.
Existing domain-specific frameworks for training GNNs, such as DGL~\cite{dgl} and
PyTorch-Geometric~\cite{pyg}, cannot scale to giant graphs. They were
mainly developed for training on a single machine. Although there have been
some efforts in building systems for distributed GNN training, they either focus
on full batch training by partitioning graphs to fit the aggregated memory
of multiple devices~\cite{roc,neugraph,tripathy2020reducing} or suffer
from the huge network traffic caused by fetching neighbor node data~\cite{aligraph,agl,euler}.
System architectures~\cite{ps,project_adam,byteps} proposed
for training neural networks for computer vision and natural language
processing are not directly applicable because one critical bottleneck in
GNN training is the network traffic of fetching neighbor node data
due to the vertex dependencies, while previous systems majorly focuses on network
traffic from exchanging the gradients of model parameters.


In this work, we develop \sys on top of DGL to perform efficient and scalable
mini-batch GNN training on a cluster of machines. It provides distributed components
with APIs compatible to DGL's existing ones. As such, it requires trivial
effort to port DGL's training code to \sys. Internally, it deploys
multiple optimizations to speed up computation. It distributes graph data
(both graph structure and the associated data, such as node and edge features)
across all machines and run trainers, sampling servers (for sampling
subgraphs to generate mini-batches) and in-memory KVStore servers (for serving node data
and edge data) all on the same set of machines. To achieve good model accuracy,
\sys follows a synchronous training approach and allows ego-networks forming
the mini-batches to include non-local nodes. To reduce network communication,
\sys adopts METIS \cite{metis} to partition a graph with minimum edge cut
and co-locate data with training computation. In addition, \sys deploys multiple
load balancing optimizations to
tackle the imbalance issue, including multi-constraint partitioning and two-level
workload splitting. \sys further reduces network communication in sampling by
replicating halo nodes in the partitioned graph structure but does not replicate
data in halo nodes to have a small memory footprint. \sys provides distributed
embeddings with efficient sparse updates for transductive graph models.

We conduct comprehensive experiments to evaluate the efficiency of \sys and
effectiveness of the optimizations. Overall, \sys achieves $2.2\times$ speedup
over Euler on a cluster of four CPU machines. The main performance advantage
comes from the efficient feature copy with $5\times$ data copy throughput.
\sys speeds up the training linearly without compromising model accuracy
as the number of machines
increases in a cluster of 16 machines and easily scales the GraphSage model
to a graph with
100 million nodes and 3 billion edges. It takes 13 seconds per epoch to train
on such a graph in a cluster of 16 machines.

\section{Background}

\subsection{Graph Neural Networks} \label{sec:gnn}

GNNs emerge as a family of neural networks capable of learning a joint representation
from both the graph structure and vertex/edge features.
Recent studies~\cite{gnn-chemistry, graphnets} formulate GNN models with
\emph{message passing}, in which vertices broadcast messages to their neighbors and compute
their own representation by aggregating received messages.

More formally, given a graph $\mathcal{G(\mathcal{V},\mathcal{E})}$, we denote
the input feature of vertex $v$ as $\mathbf{h}_v^{(0)}$, and the feature of the edge
between vertex $u$ and $v$ as $\mathbf{e}_{uv}$.
To get the representation of a vertex at layer $l$, a GNN model performs the computations
below:


\begin{equation}\label{eq:mp-vertex}
\mathbf{h}_v^{(l+1)} = g(\mathbf{h}_v^{(l)},\bigoplus_{u\in\mathcal{N}(v)} f(\mathbf{h}_u^{(l)}, \mathbf{h}_v^{(l)}, \mathbf{e}_{uv}))
\end{equation}


\noindent Here $f$, $\bigoplus$ and $g$ are customizable or parameterized functions
(e.g., neural network modules) for calculating messages, aggregating messages, and
updating vertex representations, respectively.
Similar to convolutional neural networks (CNNs), a GNN model iteratively applies
Equations~\eqref{eq:mp-vertex} to generate vertex representations for multiple layers.

There are potentially two types of model parameters in graph neural networks. $f$,
$\bigoplus$ and $g$ can contain model parameters, which are shared among all vertices.
These model parameters are updated in every mini-batch and we refer to these parameters
as \textit{dense} parameters. Some GNN models may additionally learn an \textit{embedding} for each vertex.
Embeddings are part of the model parameters and only a subset of vertex embeddings
are updated in a mini-batch. We refer to these model parameters as \textit{sparse}
parameters.

\subsection{Mini-batch training} \label{sec:minibatch}

GNN models on a large dataset can be trained in a mini-batch fashion
just like deep neural networks in other domains like computer vision and natural
language processing. However, GNN mini-batch training is different from other neural
networks due to the data dependency between vertices. Therefore, we need to carefully
sample subgraphs that capture the data dependencies in the original graph to train
GNN models.

\begin{figure}
\centering
\begin{subfigure}{.5\textwidth}
  \centering
  \includegraphics[width=.5\linewidth]{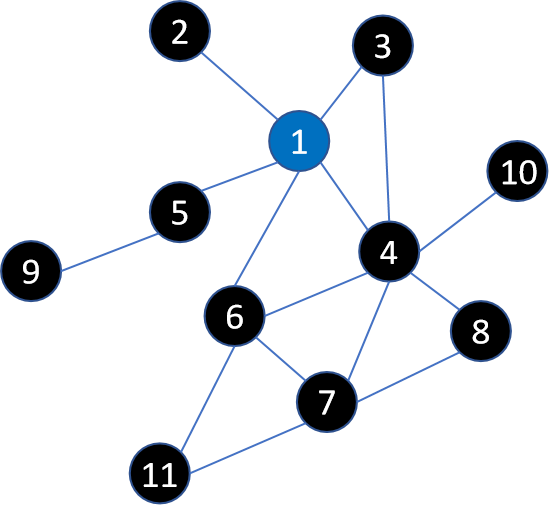}
  \caption{An input graph.}
  \label{fig:full_graph}
\end{subfigure}%
\hfill
\begin{subfigure}{.5\textwidth}
  \centering
  \includegraphics[width=.3\linewidth]{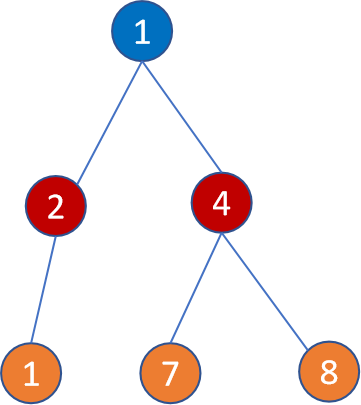}
  \caption{A sampled graph for computing one target vertex representation with a two-layer
  GNN model. Messages flow from leaves to root.}
  \label{fig:mini_batch}
\end{subfigure}
\caption{One sampled mini-batch in GNN training.}
\label{fig:test}
\end{figure}

A typical strategy of training a GNN model \cite{graphsage} follows three steps:
(i) sample a set of $N$ vertices, called \textit{target vertices}, uniformly at random from
the training set; (ii) randomly pick
at most $K$ (called \textit{fan-out}) neighbor vertices for each target vertex;
(iii) compute the target vertex representations
by gathering messages from the sampled neighbors.
When the GNN has multiple layers, the sampling is repeated recursively.
That is, from a sampled neighbor vertex, it continues sampling
its neighbors. The number of recursions is determined by the number of layers
in a GNN model. This sampling strategy forms a computation graph for
passing messages on.
Figure \ref{fig:mini_batch} depicts such a graph for computing representation of
one target vertex when the GNN has two layers. The sampled graph and together with
the extracted features are called a mini-batch in GNN training.

There have been many works regarding to the different strategies to sample graphs
for mini-batch training \cite{chen2017stochastic, fastgcn, ladies, huang2018adaptive, clustergcn}. Therefore, a GNN framework needs to be flexible as well as scalable to giant graphs.


\section{\sys System Design}

\subsection{Distributed Training Architecture}

\begin{figure}
\centering
\includegraphics[width=0.95\linewidth]{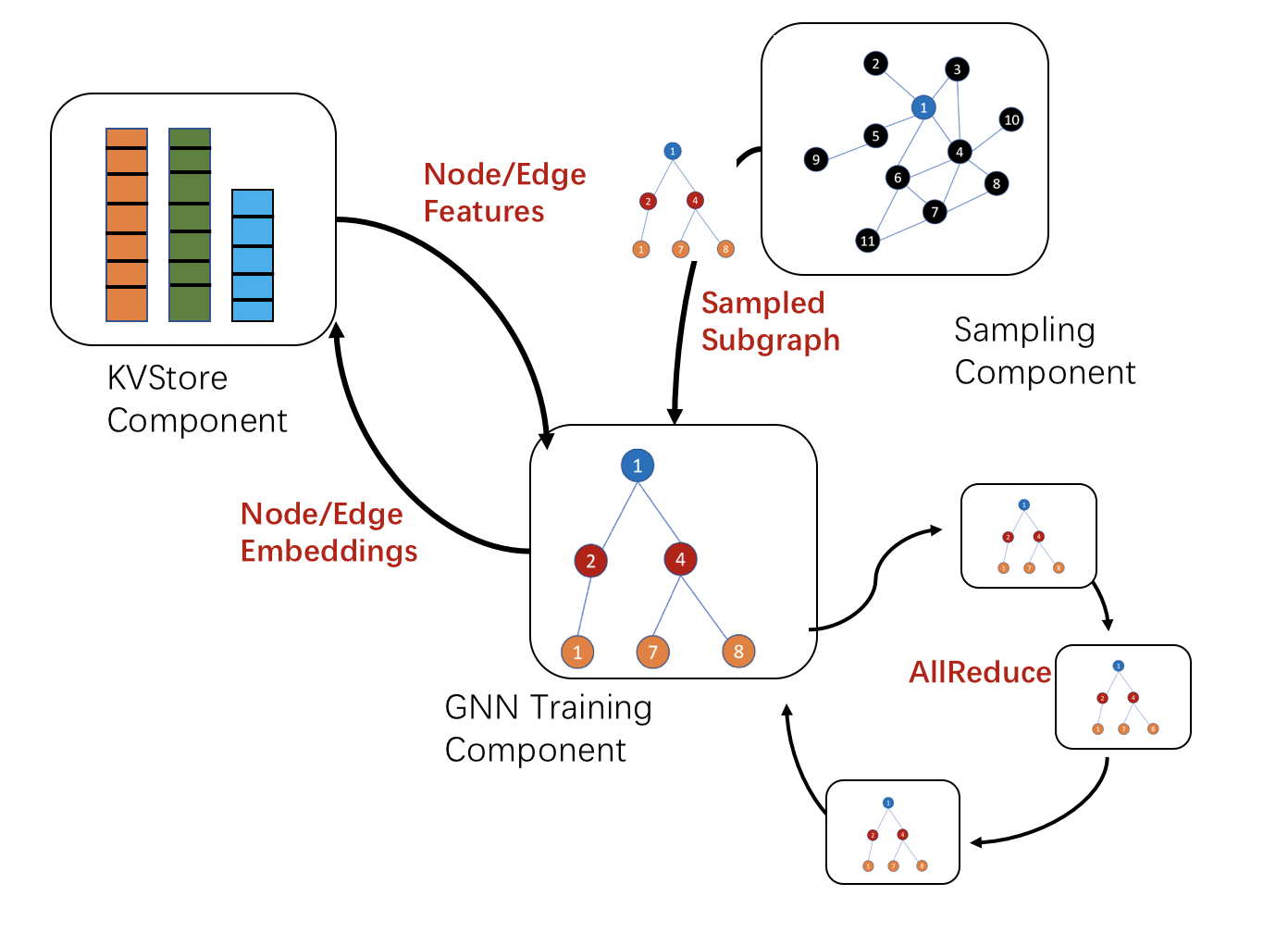}
\caption{\sys's logical components.}
\label{fig:dist}
\vspace{-1em}
\end{figure}

\sys distributes the mini-batch training process of GNN models to a cluster of machines.
It follows the synchronous stochastic gradient descent (SGD) training; each machine computes
model gradients with respect to its own mini-batch, synchronizes gradients with others
and updates the local model replica. At a high level, \sys consists of the following logical 
components (Figure~\ref{fig:dist}):

\begin{itemize}
    \item A number of \textit{samplers} in charge of sampling the mini-batch graph structures
    from the input graph. Users invoke \sys samplers in the trainer process
    via the same interface in DGL for neighbor sampling, which internally
    becomes a remote process call (RPC).
    After mini-batch graphs are generated, they are sent back to the trainers.
    \item A \textit{KVStore} that stores all vertex data and edge data distributedly.
    It provides two convenient interfaces for pulling the data from or pushing the data to
    the distributed store.
    It also manages the vertex embeddings if specified by the user-defined GNN model.
    \item A number of \textit{trainers} that compute the gradients of the model parameters
    over a mini-batch. At each iteration, they first fetch the mini-batch graphs from the
    samplers and the corresponding vertex/edge features from the KVStore. They then run the
    forward and backward computation on their own mini-batches in parallel to compute the
    gradients. The gradients of dense parameters are dispatched to the
    \textit{dense model update component} for synchronization,
    while the gradients of sparse embeddings are sent back to the KVStore for update.
    \item A \textit{dense model update component} for aggregating dense GNN parameters
    to perform synchronous SGD. \sys reuses the existing components depending on DGL's
    backend deep learning frameworks (e.g.,
    PyTorch, MXNet and TensorFlow). For example, \sys calls the all-reduce primitive
    when the backend framework is PyTorch~\cite{li2020pytorch}, or resorts to parameter servers~\cite{ps} for MXNet and TensorFlow backends.
\end{itemize}

\begin{figure}
\centering
\includegraphics[width=0.95\linewidth]{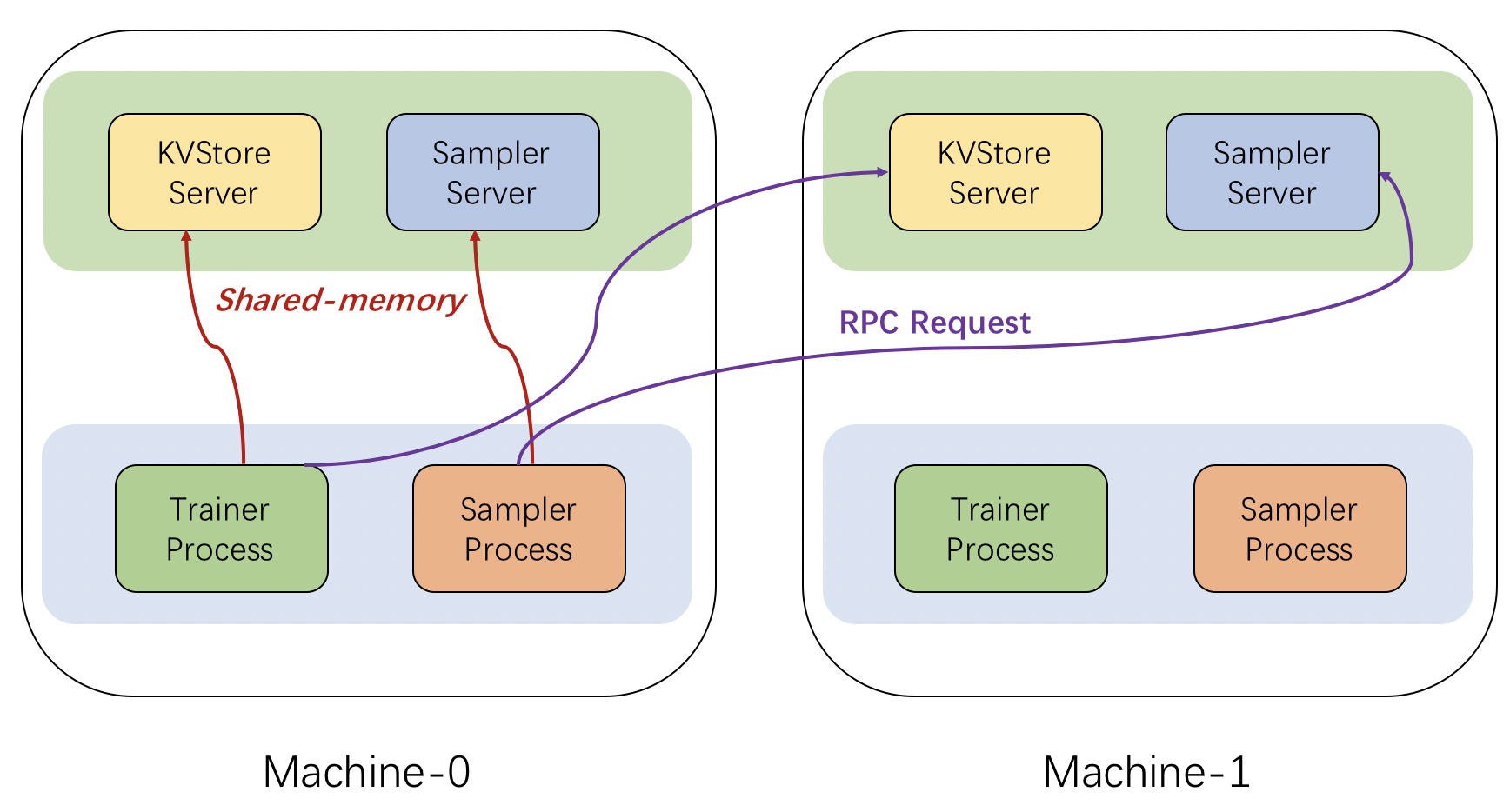}
\caption{The deployment of \sys's logical components on a cluster of two machines.}
\label{fig:arch}
\vspace{-1em}
\end{figure}

When deploying these logical components to actual hardware, the first consideration
is to reduce the network traffic among machines because graph computation is
data intensive~\cite{Eyerman2018}. \sys adopts the owner-compute rule (Figure~\ref{fig:arch}).
The general principle is to dispatch computation to the data owner to reduce
network communication. \sys first partitions the input graph with a light-weight min-cut
graph partitioning algorithm. It then partitions the vertex/edge features and co-locates
them with graph partitions. \sys launches the sampler and KVStore servers on each machine
to serve the local partition data. Trainers also run on the same cluster of machines and each
trainer is responsible for the training samples from the local partition. This design
leverages data locality to its maximum. Each trainer works on samples from the local partition
so the mini-batch graphs will contain mostly local vertices and edges. Most of the
mini-batch features are
locally available too via shared memory, reducing the network traffic significantly.
In the following sections, we will elaborate more on the design of each components.


\subsection{Graph Partitioning} \label{sec:partition}

The goal of graph partitioning is to split the input graph to multiple partitions
with a minimal number of edges across partitions.
Graph partitioning is a preprocessing step before distributed training. A graph is
partitioned once and used for many distributed training runs, so its overhead
is amortized.

\begin{figure}
\centering
\begin{subfigure}{.5\textwidth}
  \centering
  \includegraphics[width=.5\linewidth]{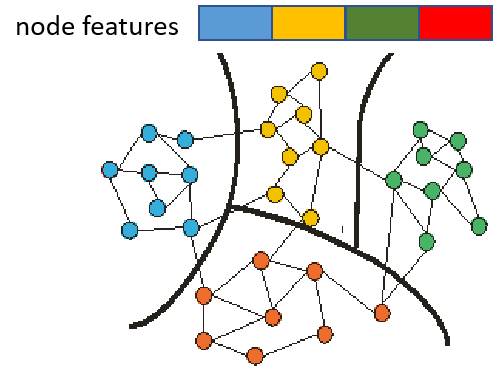}
  \caption{Assign vertices to graph partitions}
  \label{fig:assign_vertices}
\end{subfigure}%
\hfill
\begin{subfigure}{.5\textwidth}
  \centering
  \includegraphics[width=.8\linewidth]{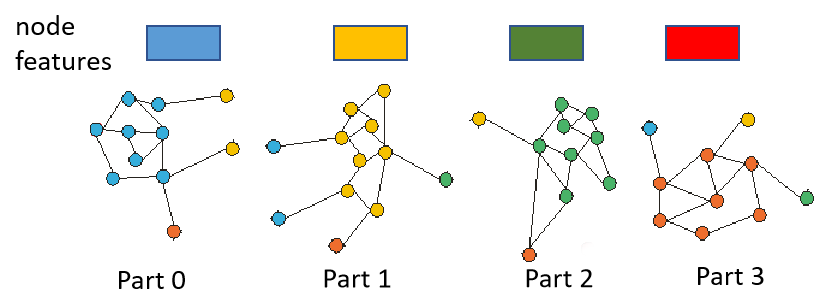}
  \caption{Generate graph partitions with HALO vertices (the vertices with
  different colors from majority of the vertices in the partition).}
  \label{fig:graph_part}
\end{subfigure}
\caption{Graph partitioning with METIS in \sys.}
\label{fig:test}
\end{figure}

\sys adopts METIS \cite{metis} to partition a graph. This algorithm assigns densely
connected vertices to the same partition to reduce the number of edge cuts between
partitions (Figure \ref{fig:assign_vertices}). After assigning some vertices to a partition,
\sys assigns all incident edges of these vertices to the same partition.
This ensures that all the neighbors of the local vertices are accessible on the partition
so that samplers can compute locally without communicating to each other.
With this partitioning strategy, each edge has a unique assignment while some
vertices may be duplicated (Figure \ref{fig:graph_part}).
We refer to the vertices assigned by METIS to a partition
as \textit{core vertices} and the vertices duplicated by our edge assignment strategy
as \textit{HALO vertices}. All the core vertices also have unique partition assignments.

While minimizing edge cut, \sys deploys multiple strategies to balance
the partitions so that mini-batches of different trainers are roughly balanced.
By default, METIS only balances the number of
vertices in a graph. This is insufficient to generate balanced partitions
for synchronous mini-batch training, which requires the same number of
batches from each partition per epoch and
all batches to have roughly the same size. We formulate this load balancing
problem as a multi-constraint partitioning problem, which balances
the partitions based on user-defined constraints~\cite{multiconstraint}.
\sys takes advantage of the multi-constraint mechanism in METIS to
balance training/validation/test vertices/edges in
each partition as well as balancing the vertices of different types and
the edges incident to the vertices of different types.

METIS' partitioning algorithms are based on the multilevel paradigm,
which has been shown to produce high-quality partitions. However, for
many types of graphs involved in learning on graphs tasks
(e.g., graphs with power-law degree distribution), the successively
coarser graphs become progressively denser, which considerably increases
the memory and computational complexity of multilevel algorithms.
To address this problem, we extended METIS to only retain a subset of 
the edges in each successive graph so that the degree of each coarse vertex
is the average degree of its constituent vertices. This ensures that as
the number of vertices in the graph reduces by approximately a factor of two,
so do the edges. To ensure that the partitioning solutions obtained in
the coarser graphs represent high-quality solutions in the finer graphs,
we only retain the edges with the highest weights in the coarser graph.
In addition, to further reduce the memory requirements, we use an out-of-core strategy
for the coarser/finer graphs that are not being processed currently.
Finally, we run METIS by performing a single initial partitioning (default is 5) 
and a single refinement iteration (default is 10) during each level.
For power-law degree graphs, this optimization leads to a small increase in 
the edge-cut (2\%-10\%) but considerably reduces its runtime. Overall,
the set of optimizations above compute high-quality partitionings
requiring ~5$\times$ less memory and ~8$\times$ less time than METIS'
default algorithms.

After partitioning the graph structure, we also partition vertex features and
edge features based on the graph partitions. We only assign the features of
the \textit{core vertices} and edges of a partition to the partition. Therefore,
the vertex features and edge features are not duplicated.

After graph partitioning, \sys manages two sets of vertex IDs and edge IDs.
\sys exposes global vertex IDs and edge IDs for model developers to identify
vertices and edges. Internally,
\sys uses local vertex IDs and edge IDs to locate vertices and edges in a partition
efficiently, which is essential to achieve high system speed as demonstrated
by previous works \cite{gemini}. To save memory for maintaining
the mapping between global IDs and local IDs, \sys relabels vertex IDs
and edge IDs of the input graph during graph partitioning to ensure that all
IDs of core vertices and edges in a partition fall into a contiguous ID range.
In this way, mapping a global ID to a partition is binary lookup in
a very small array
and mapping a global ID to a local ID is a simple subtraction operation.

\subsection{Distributed Key-Value Store}
The features of vertices and edges are partitioned and stored in multiple machines.
Even though \sys partitions a graph to assign
densely connected vertices to a partition, we still need to read data from
remote partitions. To simplify the data access
on other machines, \sys develops a distributed in-memory key-value store (KVStore)
to manage the vertex and edge features as well as vertex embeddings, instead of
using an existing distributed in-memory KVStore, such as Reddis, for
\textit{(i)} better co-location of node/edge features in KVStore and graph partitions,
\textit{(ii)} faster network access for high-speed network, \textit{(iii)}
efficient updates on sparse embeddings.

\sys's KVStore supports flexible partition policies to map data to different machines.
For example, vertex data and edge data are usually partitioned and mapped to
machines differently as shown in Section \ref{sec:partition}. \sys defines separate
partition policies for vertex data and edge data, which aligns with the graph partitions
in each machine.

Because accessing vertex and edge features usually accounts for the majority of
communication in GNN distributed training,
it is essential to support efficient data access in KVStore. A key optimization
for fast data access is to use shared memory. Due to the co-location of data
and computation, most of data access to KVStore results in the KVStore server
on the local machine. Instead of going through Inter-Process Communication (IPC),
the KVStore server shares all data with the trainer process via shared memory.
Thus, trainers can access most of the data directly without paying any overhead
of communication and process/thread scheduling.
We also optimize network transmission of \sys's KVStore for fast networks
(e.g., 100Gbps network). We develop an optimized RPC framework for fast networking
communication, which adopts zero-copy mechanism for data serialization and
multi-thread send/receive interface.

In addition to storing the feature data, we design \sys's KVStore to
support sparse embedding for training transductive models with
learnable vertex embeddings. Examples are knowledge graph embedding
models \cite{dglke}. In GNN mini-batch training, only a small subset of
vertex embeddings are involved in the computation and updated during each iteration.
Although almost all deep learning frameworks have off-the-shelf sparse embedding modules,
most of them lack efficient support of distributed sparse update.
\sys's KVStore shards the vertex embeddings in the same way as vertex features.
Upon receiving the embedding gradients (via the PUSH interface),
KVStore updates the embedding based on the optimizer the user registered.


\subsection{Distributed Sampler} \label{sec:sampling}

DGL has provided a set of flexible Python APIs to support a variety
of sampling algorithms proposed in the literature.
\sys keeps this API design but with a different internal implementation.
At the beginning of each iteration, the trainer
issues sampling requests using the target vertices in the current mini-batch.
The requests are dispatched to the machines according to the core vertex assignment
produced by the graph partitioning algorithm. Upon receiving the request,
sampler servers call DGL's sampling operators on the local partition and
transmit the result back to the trainer process. Finally, the trainer collects
the results and stitches them together to generate a mini-batch.

\sys deploys multiple optimizations to effectively accelerate mini-batch generation.
\sys can create multiple sampling worker processes for each trainer to sample mini-batches
in parallel. By issuing sampling requests to the sampling workers, trainers overlap
the sampling cost with mini-batch training.
When a sampling request goes to the local sampler server,
the sampling workers to access the graph structure stored
on the local sampler server directly via shared memory to avoid the cost of the RPC stack.
The sampling workers
also overlaps the remote RPCs with local sampling computation by first issuing remote
requests asynchronously. This effectively hides the network latency because the
local sampling usually accounts for most of the sampling time.
When a sampler server receives sampling requests, it only needs to sample vertices
and edges from the local partition because
our graph partitioning strategy (Section~\ref{sec:partition})
guarantees that the core vertices in a partition have the access to the entire
neighborhood.



\subsection{Mini-batch Trainer}
Mini-batch trainers run on each machine to jointly estimate gradients and update
parameters of users' models. \sys provides utility functions to split the training
set distributedly and generate balanced workloads between trainers.

Each trainer samples data points uniformly at random to generate mini-batches
independently.
Because DistDGL generates balanced partitions (each partition has roughly the same
number of nodes and edges) and uses synchronous SGD to train
the model, the data points sampled collectively by all trainers in each iteration
are still sampled uniformly at random across the entire dataset. As such,
distributed training in DistDGL in theory does not affect the convergence rate
or the model accuracy.

\begin{figure}
\centering
\includegraphics[width=0.5\linewidth]{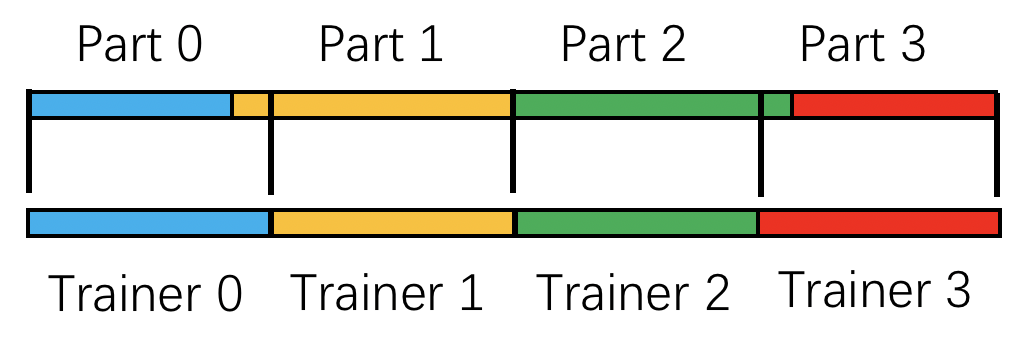}
\caption{Split the workloads evenly to balance the computation among trainer processes.}
\label{fig:split}
\vspace{-1em}
\end{figure}

To balance the computation in each trainer, \sys uses a two-level strategy to split
the training set evenly across all trainers at the beginning of distributed training.
We first ensure that each trainer has the same number of training samples.
The multi-constraint algorithm
in METIS (Section \ref{sec:partition}) can only assign roughly the same number of
training samples (vertices or edges) to each partition (as shown by the rectangular boxes
on the top in Figure \ref{fig:split}). We thus evenly split the training samples
based on their IDs and assign the ID range
to a machine whose graph partition has the largest overlap with the ID range.
This is possible because we relabel vertex and edge IDs during graph partitioning
and the vertices and edges in a partition have a contiguous ID range.
There is a small misalignment between the training samples assigned to a trainer and
the ones that reside in a partition. Essentially, we make a tradeoff between
load balancing and data locality. In practice, as long as the graph partition algorithm 
balances the number of training samples between partitions, the tradeoff is negligible.
If there are multiple trainers on one partition,
we further split the local training vertices evenly and assign
them to the trainers in the local machine.
We find that random split in practice gives a fairly balanced workload assignment.

In terms of parameter synchronization, we use synchronous SGD to update dense model parameters.
Synchronous SGD is commonly used to train deep neural network models and usually
leads to better model accuracy.
We use asynchronous SGD to update the sparse vertex embeddings
in the Hogwild fashion \cite{hogwild} to overlap communication and computation.
In a large graph, there are many vertex embeddings. Asynchronous SGD updates
some of the embeddings in a mini-batch. Concurrent updates from multiple trainers
rarely result in conflicts because mini-batches from different trainers run on
different embeddings.
Previous study~\cite{dglke} has verified that asynchronous update of sparse embeddings can
significantly speed up the training with nearly no accuracy loss.

For distributed CPU training, \sys parallelizes the computation with both multiprocessing
and multithreading. Inside a trainer process, we use OpenMP to parallelize
the framework operator computation (e.g., sparse matrix multiplication and
dense matrix multiplication).
We run multiple trainer processes on each machine to parallelize the computation
for non-uniform memory architecture (NUMA), which is a typical architecture for
large CPU machines. This hybrid approach is potentially more advantageous than
the multiprocessing approach for synchronous SGD because we need to aggregate
gradients of model parameters from all trainer processes and broadcast new model
parameters to all trainers. More trainer processes result in more communication
overhead for model parameter updates.

\section{Evaluation}

In this section, we evaluate \sys to answer the following questions:
\begin{itemize}
    \item \emph{Can \sys train GNNs on large-scale graphs and accelerate the training with more machines?}
    \item \emph{Can \sys's techniques effectively increase the data locality for GNN training?}
    \item \emph{Can our load balancing strategies effectively balance the workloads in the cluster of machines?}
\end{itemize}

\begin{table}
    \centering
    \caption{Dataset statistics from the Open Graph Benchmark~\cite{ogb}.}\label{tbl:dataset}
    \begin{tabular}{lrrr}\toprule
         Dataset & \# Nodes & \# Edges & Node Features \\\midrule
         \textsc{ogbn-product} & 2,449,029 & 61,859,140 & 100 \\
         \textsc{ogbn-papers100M} & 111,059,956 &  3,231,371,744 & 128 \\
         \bottomrule
    \end{tabular}
\end{table}

We focused on the node classification task using GNNs throughout the evaluation. The GNNs for other tasks such as link prediction mostly differ in the objective function while sharing most of the GNN architectures so we omit them in the experiments.

We benchmark the state-of-the-art GraphSAGE~\cite{graphsage} model on two
Open Graph Benchmark (OGB) datasets \cite{ogb} shown in Table \ref{tbl:dataset}.
The GraphSAGE model has three layers of hidden size 256; the sampling fan-outs
of each layer are 15, 10 and 5. We use a cluster of eight AWS EC2 m5n.24xlarge
instances (96 VCPU, 384GB RAM each) connected by a 100Gbps network.

In all experiments, we use DGL v0.5 and Pytorch 1.5. For Euler experiments,
we use Euler v2.0 and TensorFlow 1.12.

\subsection{\sys vs. other distributed GNN frameworks}

We compare the training speed of \sys with Euler \cite{euler}, one of
the state-of-the-art distributed GNN training frameworks, on four m5n.24xlarge
instances. Euler is designed
for distributed mini-batch training, but it adopts different parallelization
strategy from \sys. It parallelizes computation completely with multiprocessing
and uses one thread for both forward and backward computation as well as
sampling inside a trainer. To have a fair comparison between the two frameworks,
we run mini-batch training with the same global batch size (the total size of
the batches of all trainers in an iteration) on both frameworks because we
use synchronized SGD to train models.

\begin{figure}
\centering
\begin{subfigure}{.5\textwidth}
  \centering
  \includegraphics[width=.9\linewidth]{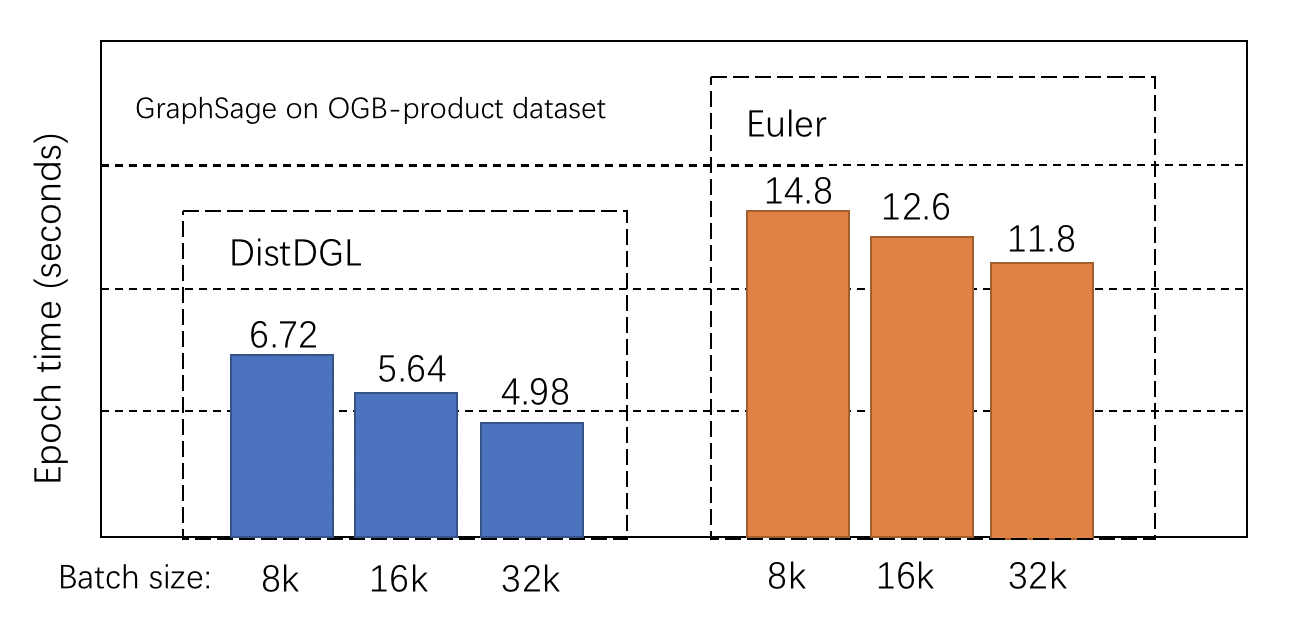}
  \caption{The overall runtime per epoch with different global batch sizes.}
  \label{fig:overall_runtime}
\end{subfigure}%
\hfill
\begin{subfigure}{.5\textwidth}
  \centering
  \includegraphics[width=.95\linewidth]{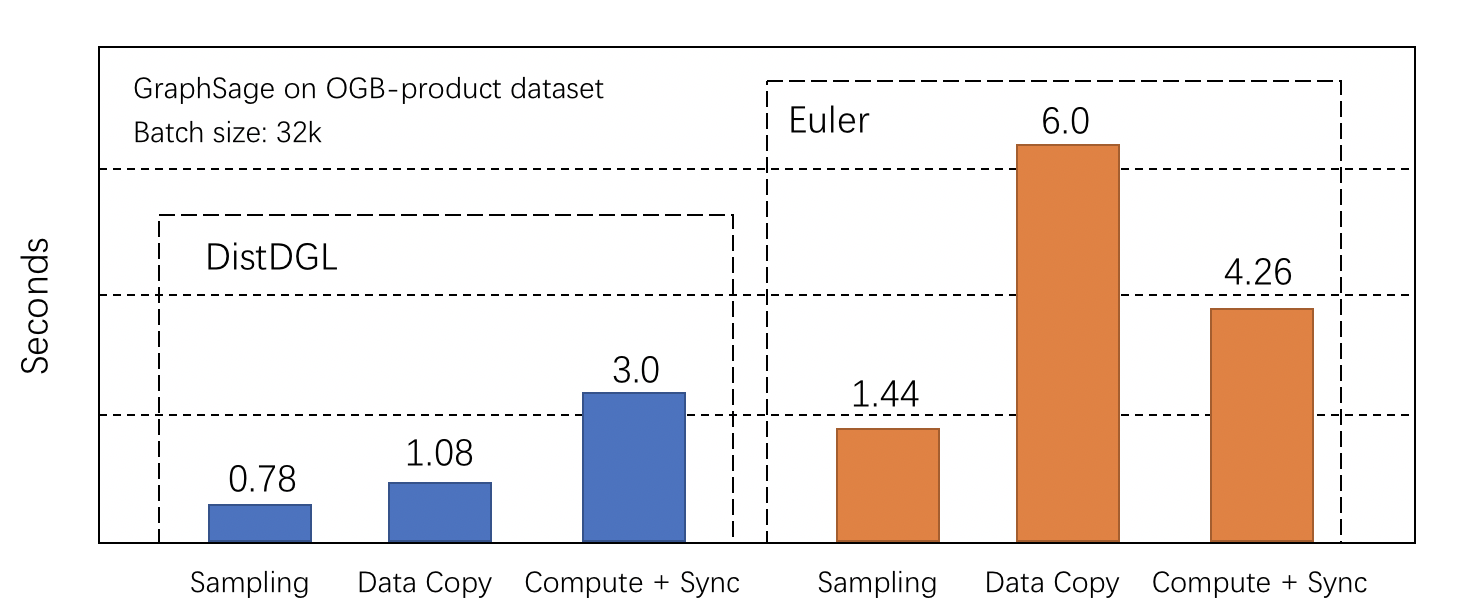}
  \caption{The breakdown of epoch runtime for the batch size of 32K.}
  \label{fig:breakdown}
\end{subfigure}
\caption{\sys vs Euler on \textsc{ogbn-product} graph on four m5n.24xlarge instances.}
\label{fig:dgl_vs_euler}
\end{figure}

\sys gets $2.2\times$ speedup over Euler in all different batch sizes
(Figure \ref{fig:overall_runtime}). To have a better understanding of
\sys's performance advantage, we break down the runtime of
each component within an iteration shown in Figure \ref{fig:breakdown}.
The main advantage of \sys is \textit{data copy}, in which \sys has more than
$5\times$ speedup. This is expected because
\sys uses METIS to generate partitions with minimal edge cuts and trainers
are co-located with the partition data to reduce network communication.
The speed of \textit{data copy} in \sys gets close to local memory copy
while Euler has to copy data through TCP/IP from the network. \sys also
has $2\times$ speedup in \textit{sampling} over Euler for the same reason:
\sys samples majority of vertices and edges from the local partition
to generate mini-batches. \sys relies on DGL and Pytorch to perform
sparse and dense tensor computation in a mini-batch and uses Pytorch to synchronize
gradients among trainers while Euler relies on TensorFlow for both mini-batch
computation and gradient synchronization. \sys is slightly faster
in mini-batch computation and gradient synchronization. Unfortunately,
we cannot separate the batch computation and gradient synchronization
in Pytorch.

\subsection{\sys's sparse embedding vs. Pytorch's sparse embedding}
Many graph datasets do not have vertex features. We typically use transductive
GNN models with learnable vertex embeddings for these graphs.
\sys provides distributed embeddings for such use case, with optimizations
for sparse updates. Deep learning frameworks, such as
Pytorch, also provide the sparse embedding layer for similar use cases and
the embedding layer can be trained in a distributed fashion. To evaluate
the efficiency of \sys's distributed embeddings, we adapt the GraphSage
model by replacing vertex data of the input graph with \sys's
or Pytorch's sparse embeddings.

\begin{figure}
\centering
\includegraphics[width=0.95\linewidth]{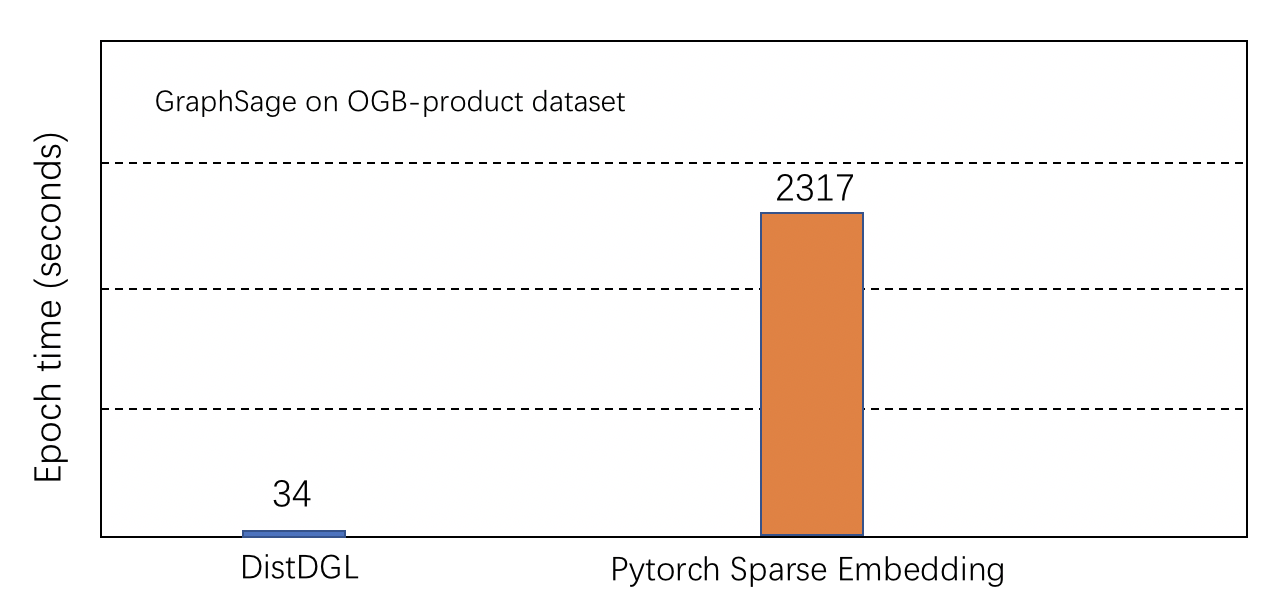}
\caption{The GraphSage model with \sys's and Pyotch's sparse Embedding on
the \textsc{ogbn-product} graph.}
\label{fig:dgl-vs-pytorch}
\end{figure}

The GraphSage model with \sys's sparse embeddings on \textsc{ogbn-product} graph
gets almsot $70\times$ speedup over the version with Pytorch sparse embeddings
(Figure \ref{fig:dgl-vs-pytorch}). The main difference is that \sys's sparse
embeddings are updated with \sys's efficient KVStore, which is natural for
implementing sparse embedding updates. As such, it gets all benefits of \sys's
optimizations, such as co-location of data and computation.
In contrast, Pytorch's sparse embeddings
are updated with its DistributedDataParallel module. Essentially, it is implemented
with the AllReduce primitive, which requires the gradient tensor exchanged
between trainers to have exactly the same shape. As such, Pytorch has to pad
the gradient tensor of sparse embeddings to the same size.

\subsection{Scalability}
We further evaluate the scalability of \sys in the EC2 cluster. In this experiment,
we fix the mini-batch size in each trainer and increase the number of trainers
when the number of machines increases. We use the batch size of 2000 per trainer.

\begin{figure}
\centering
\includegraphics[width=0.95\linewidth]{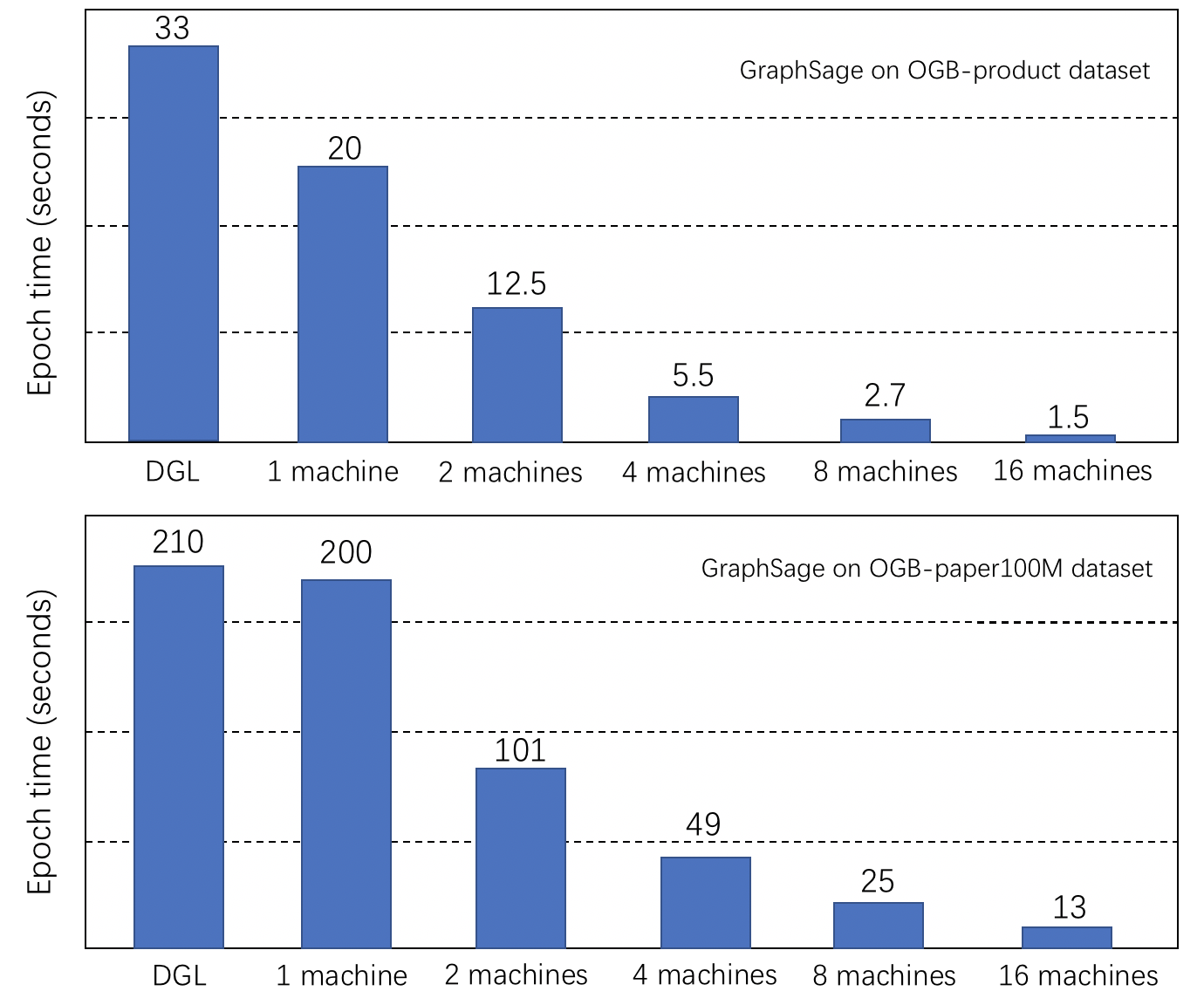}
\caption{\sys achieves linear speedup w.r.t. the number of machines.
}
\label{fig:scalability}
\end{figure}

\sys achieves a linear speedup as the number of machines increases in the cluster
(Figure \ref{fig:scalability}) for both OGB datasets. When running on a larger
cluster, \sys needs to perform more sampling on remote machines and fetch more data
from remote machines. This linear speedup indicates that our optimizations
prevent network communication from being the bottleneck. It also suggests that
the system is well balanced when the number of machines increases. With all of
our optimizations, \sys can easily scale to large graphs with hundreds of millions
of nodes. It takes only 13 seconds to train the GraphSage model on
the \textsc{ogbn-papers100M} graph in a cluster of 16 m5.24xlarge machines.

We also compare
\sys with DGL's multiprocessing training (two trainer processes).
\sys running on a single machine with two trainers outperforms DGL.
This may attribute to the different multiprocessing sampling used by
the two frameworks. DGL relies on Pytroch dataloader's multiprocessing
to sample mini-batches while \sys uses dedicated sampler
processes to generate mini-batches.

\begin{figure}
\centering
\includegraphics[width=0.95\linewidth]{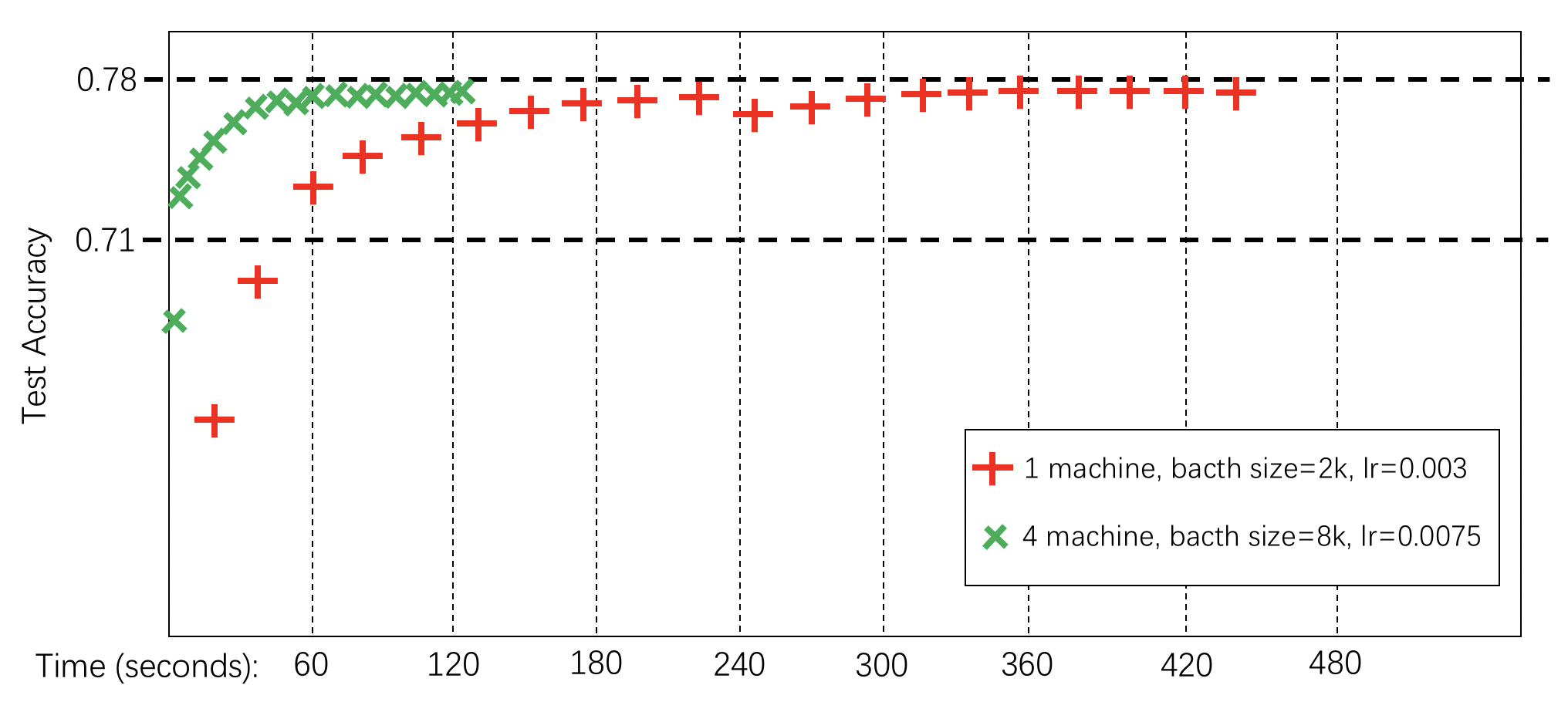}
\caption{\sys convergence of distributed training.
}
\label{fig:convergence}
\end{figure}

In addition to the training speed, we also verify the training accuracy
of \sys on different numbers of machines (Figure \ref{fig:convergence}).
We can see that \sys quickly converges to almost the same peak accuracy
achieved by the single-machine training, which takes a much longer time
to converge.



\subsection{Ablation Study}
We further study the effectiveness of the main optimizations in \sys: 1) reducing
network traffic by METIS graph partitioning and co-locating data and computation,
2) balance the graph partitions with multi-constraint partitioning.
To evaluate their effectiveness, we compare \sys's graph partitioning algorithm
with two alternatives: random graph partitioning and default METIS partitioning
without multi-constraints. We use a cluster of four machines to run the experiments.

\begin{figure}
\centering
\includegraphics[width=0.95\linewidth]{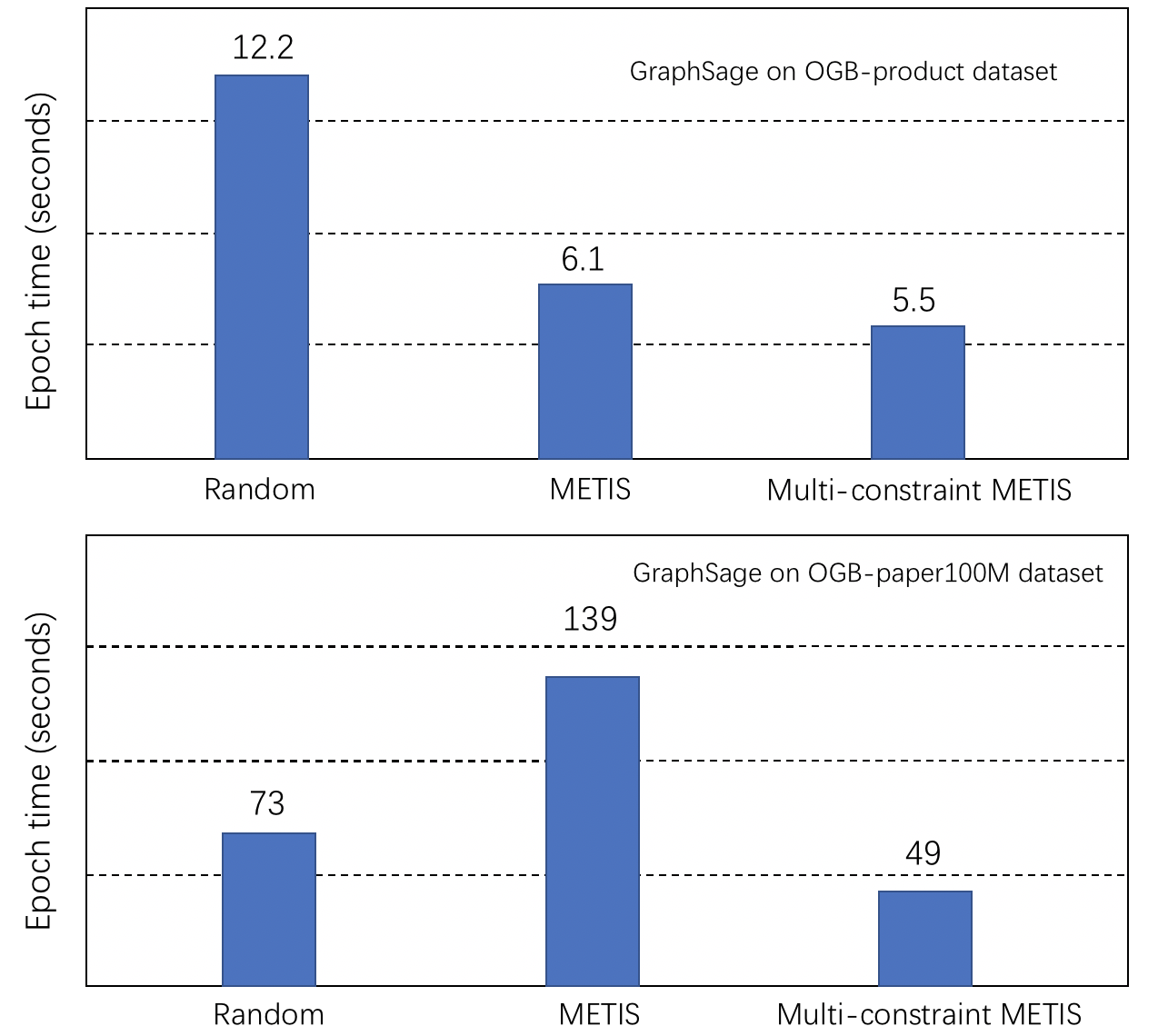}
\caption{METIS vs Random Partition on four machines}
\label{fig:metis}
\end{figure}

METIS partitioning with multi-constraints to balance the partitions achieves
good performance on both datasets (Figure \ref{fig:metis}). Default METIS partitioning
performs well compared with random partitioning ($2.14\times$ speedup) on
the \textsc{ogbn-product} graph due to
its superior reduction of network communication; adding multiple constraints to
balance partitions gives additional 4\% improvement over default METIS partitioning.
However, default METIS partitioning achieves much worse performance than random
partitioning on the \textsc{ogbn-papers100M} graph due to high imbalance between partitions
created by METIS, even though METIS can effectively reduce the number of edge
cuts between partitions. Adding multi-constraint optimizations to balance
the partitions, we see the benefit of reducing network communication. This
suggests that achieving load balancing is as important as reducing network
communication for improving performance.
\section{Related Work}

\subsection{Distributed DNN Training}
There are many system-related works to optimize distributed deep neural
network (DNN) training. The parameter server~\cite{li2014scaling} is designed
to maintain and update the sparse model parameters. Horovod~\cite{horovod}
and Pytorch distributed~\cite{li2020pytorch}
uses allreduce to aggregate dense model parameters but does not work for
sparse model parameters. BytePs~\cite{byteps} adopts more sophisticated
techniques of overlapping model computation and gradient communication to
accelerate dense model parameter updates.
Many works reduces the amount of communication by using quantization~\cite{seide20141}
or sketching~\cite{ivkin2019communicationefficient}.
Several recent work focuses on relaxing the synchronization of weights
~\cite{ho2013more, luo2019hop} in case some workers run slower
than others temporally due to some hardware issues.
GNN models are composed of multiple operators organized into
multiple graph convolution network layers shared among all nodes and edges.
Thus, GNN training also has dense parameter updates. However, the network traffic
generated by dense parameter updates is relatively small compared with
node/edge features. Thus, reducing the network traffic of dense parameter
updates is not our main focus for distributed GNN training.

\subsection{Distributed GNN Training}
A few works have been developed to scale GNN training on large graph data
in the multi-GPU setting or distributed setting.
Some of them~\cite{roc, neugraph, tripathy2020reducing}
perform full graph training on multiple GPUs or distributed memory whose
aggregated memory fit the graph data. However, we believe full graph
training is an inefficient way to train a GNN model in a large graph data
because one model update requires significant amount of computation.
The mini-batch training has been widely adopted in training a neural
network.

Multiple GNN frameworks \cite{aligraph, agl, euler} built by industry
adopt distributed mini-batch training. However, none of these frameworks
adopt locality-aware graph partitioning and co-locate data and communication.
As shown in our experiment, reducing communication is a key to
achieve good performance.

\subsection{Distributed graph processing}
There are many works on distributed graph processing frameworks.
Pregel \cite{pregel} is one of the first
frameworks that adopt message passing and vertex-centric interface
to perform basic graph analytics algorithms such as breadth-first search
and triangle counting. PowerGraph \cite{powergraph} adopts vertex cut
for graph partitioning and gather-and-scatter interface for computation.
PowerGraph had significant performance improvement overhead Pregel.
Gemini \cite{gemini} shows that previous distributed graph processing
framework has significant overhead in a single machine. It adopts the approach
to improve graph computation in a single machine first before optimizing
for distributed computation. Even though the computation pattern of distributed
mini-batch training of GNN is very different from traditional graph analytics
algorithms, the evolution of graph processing frameworks provide valuable lessons
for us and many of the general ideas, such as locality-aware graph partitioning
and co-locating data and computation, are borrowed to optimize
distributed GNN training.
\section{Conclusion}
We develop \sys for distributed GNN training. We adopt Metis partitioning to
generate graph partitions with minimum
edge cuts and co-locate data and computation to reduce the network communication.
We deploy multiple strategies to balance the graph partitions and mini-batches
generated from each partition.
We demonstrate that achieving high training speed requires both network communication
reduction and load balancing.
Our experiments show \sys has linear speedup
of training GNN models on a cluster of CPU machines without compromising model accuracy.



\bibliographystyle{IEEEtran}
\bibliography{dgl}

\vspace{12pt}

\end{document}